\documentclass[runningheads]{llncs}
\usepackage{graphicx}
\usepackage{amsmath}
\usepackage{amssymb}
\usepackage{booktabs}
\usepackage{wrapfig,lipsum}
\usepackage{subcaption}	
\usepackage{xcolor}
\usepackage{makecell}
\usepackage{cite}

\usepackage[pagebackref,breaklinks]{hyperref}
\usepackage{float}
\usepackage{bm}

\floatstyle{plaintop}
\restylefloat{table}

\newcommand{\samelineand}{\qquad}

\begin{document}

\title{Machine Learning methods for simulating particle response in
the Zero Degree Calorimeter at the ALICE experiment, CERN}% Force line breaks with \\

\titlerunning{ }

\author{
Jan Dubiński\inst{1} \and
Kamil Deja\inst{1} \and
Sandro Wenzel\inst{2} \and \\
Przemysław Rokita \inst{1} \and
Tomasz Trzcinski\inst{1,3,4,5}}
\authorrunning{J. Dubiński et al.}

\institute{$^1$Warsaw University of Technology
\samelineand  $^2$CERN \\
\samelineand $^3$Jagiellonian University 
\samelineand $^4$Tooploox \\
\samelineand $^5$IDEAS NCBR \\
\email{jan.dubinski.dokt@pw.edu.pl}}
\maketitle    

\begin{abstract}
Currently, over 50\% of the computing power at CERN’s GRID is used to run High Energy Physics simulations. 
The recent updates at the Large Hadron Collider (LHC) create the need for developing more efficient simulation methods. 
In particular, there exist a demand for a fast simulation of the neutron Zero Degree Calorimeter, where existing Monte Carlo-based methods impose a significant computational burden.
We propose an alternative approach to the problem that leverages machine learning. 
Our solution utilises neural network classifiers and generative models to directly simulate the response of the calorimeter. 
In particular, we examine the performance of variational autoencoders and generative adversarial networks, expanding the GAN architecture by an additional regularisation network and a simple, yet effective postprocessing step.
Our approach increases the simulation speed by 2 orders of magnitude while maintaining the high fidelity of the simulation.

\end{abstract}

\section{Introduction}
\label{sec:intro}

At the European Organisation for Nuclear Research (CERN) located near Geneva, Swizterland physicists and engineers study the fundamental properties of matter through High Energy Physics (HEP) experiments. Inside the Large Hadron Collider (LHC), two particle beams are being accelerated nearly to the speed of light and brought to collide in order to recreate the extreme conditions of the early universe just after the Big Bang. 

Understanding what happens during these collisions requires complex simulations that generate the expected response of the detectors inside the LHC. The currently used methods are based on statistical Monte Carlo simulations of physical interactions of particles. The high-fidelity results they provide come at a price of high computational cost. Currently, standard simulation procedures occupy the majority of CERN's computing grid system (over 500 000 CPUs in 170 centres). To address the shortcomings of this approach, an alternative solution for simulation in high-energy physics experiments that leverages generative machine learning techniques has been proposed recently. \cite{paganini2017calogan, deja2020e2esinkhorn, dubinski2022}

In this work, we examine the performance of machine learning models on the task of simulating the data from the neutron Zero Degree Calorimeter (ZDC) from the ALICE experiment, CERN. We apply a variational autoencoder and generative adversarial networks to the problem treating the results as baselines. Moreover, we expand the GAN architecture with an additional regularisation network and a simple, yet effective postprocessing step. Our solution uses a neural network classifier to filter inputs that do not cause any response of the calorimeter before passing the data to the generative model.

The proposed models are able to generate the end data directly, without simulating the effect of every physical law and interaction between particles and the experiment's matter separately. Therefore, this approach greatly reduces the demand for computational power. Our approach increases the simulation speed by 2 orders of magnitude while maintaining the high fidelity of the simulation.

\section{Related work}
\label{sec:related_work}

The need for simulating complex processes exists across many scientific domains. In recent years, solutions based on generative machine learning models have been proposed as an alternative to existing methods in cosmology~\cite{10.1093/gigascience/giab005} and genetics~\cite{Rodr_guez_2018}.
However, one of the most profound applications for generative simulations is in the field of High Energy Physics, where machine learning models can be used as a resource-efficient alternative to classic Monte Carlo-based\cite{incerti2018geant4} approaches.

Recent attempts to simulate High Energy Physics experiments ~\cite{paganini2017calogan, kansal2021particle, Erdmann_2019} leverage solutions based on Generative Adversarial Networks~\cite{goodfellow2014generative} or Variational Autoencoders ~\cite{kingma2013auto}. To the best of our knowledge ~\cite{paganini2017calogan} is the first attempt to simulate a CERN calorimeter with generative machine learning models. The authors combine three parallel GAN processing streams and an attention mechanism to simulate the response of an
electromagnetic calorimeter. The authors of ~\cite{Chekalina_2019} use Wasserstein GAN to simulate the response of another electromagnetic calorimeter. Similarly to our method, the authors embed a regressor pretrained on predicting input particle parameters in the model. This network extension allows them to overcome the difficulties with conditioning on continuous values.
In \cite{Buhmann_2021} the authors investigate the use of a network architecture dubbed Bounded Information Bottleneck Autoencoder to simulate an electromagnetic calorimeter. Their approach employs multiple additional regularization networks. Additionally, this work utilizes a post-processing network which must be trained jointly with the remaining network components. Our method employs a similar post-processing step, however, it does not require the training of an additional neural network.

\section{Zero Degree Calorimeter simulation}
\label{sec:ZDC}

The neutron Zero Degree CalorimeterN is a quartz-fiber spaghetti calorimeter, which will measure the energy of the spectator neutrons in heavy ion collisions at the CERN LHC. Its principle of operation is based on the detection of the Cherenkov light produced by the charged particles of the shower in silica optical fibres, embedded in a W-alloy absorber. \cite{ARNALDI2006235}. One out of every two fibres is sent to a photomultiplier (PMTc), while the remaining fibers are collected in bundles and sent to four different photomultipliers (PMT1 to PMT4) forming four independent towers. This segmentation allows to check the relative behaviour of the different towers and to give a rough localization of the spectator neutron’s spot on the front face of the calorimeter. The information coming from the PMTc provides a complementary measurement of the shower’s energy, in particular, useful for calibration purposes. 
Since the number of photons collected by each tower (further referred to as channels) is directly used in the further analysis of the calorimeter's output, we aim to achieve the best possible agreement, measured between the distributions of channel values for the original and fast simulation.

Simulating the response of the Zero Degree Calorimeter (ZDC) offers a challenging benchmark for generative models. The dataset consists of over 8 million samples obtained from the GEANT4 \cite{incerti2018geant4} simulation tool. Each response is created by a single particle described with 9 attributes (mass, energy, charge, momenta, primary vertex).

\begin{figure}[h!]
\centering
\includegraphics[scale = 0.24]{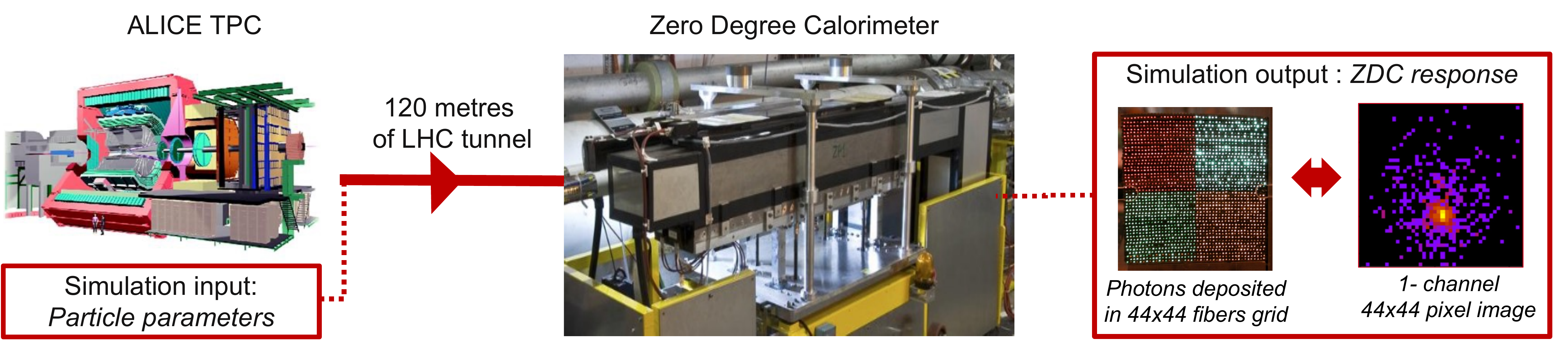}
\caption{Fast simulation of the Zero Degree Calorimeter}
\label{fig:zdc_sim}
\end{figure}

During the simulation process, the particle is propagated through the detector for over 100 meters while simulation tools must account for all of its interactions with the detector’s matter. The end result of the simulation is the energy deposited in the calorimeter’s fibres, which are arranged in a grid with 44 × 44 size.  We treat the calorimeter’s response as a 1-channel image with 44 × 44 pixels, where pixel values are the number of photons deposited in a given fibre. The schema of the simulation is depicted in Fig.\ref{fig:zdc_sim}.

To create the dataset the simulation was run multiple times for the same input particles. For that reason, multiple possible outcomes correspond to the same particle properties.  We refer to this dataset as HEP.

Importantly, over 95\% of input particles does not produce any response of the calorimeter. For that reason, there are only 300 thousand non-zero ZDC responses in the dataset. We randomly split the dataset into training (80\%) and validation (20\% subsets.

\section{Method}
\label{sec:method}
\begin{figure}[h!]
\centering
\includegraphics[scale = 0.3]{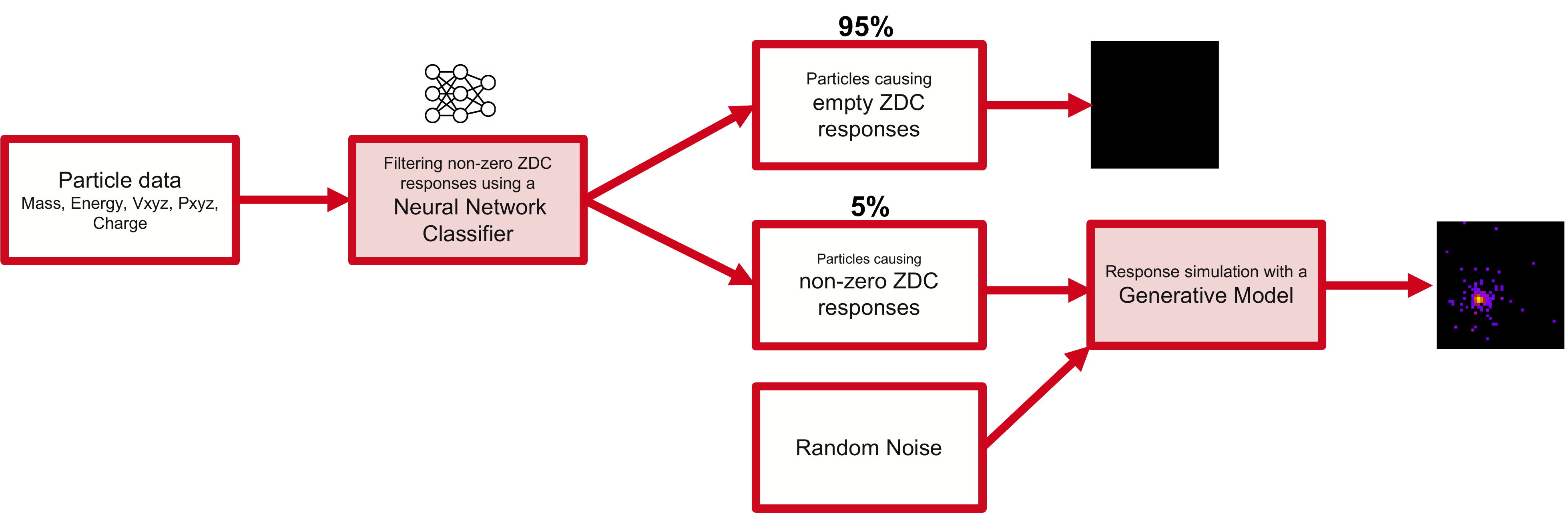}
\caption{Simulation pipeline.}
\label{fig:method}
\end{figure}

The proposed method for simulating the response of the ZDC consists of two main parts. First, we pass the input particle parameters to a binary classifier that assigns one of two possible class labels to the particle - zero or non-zero. If a particle produces an empty ZDC response, the simulation returns a 44x44 matrix of zeros. If a particle is labelled as producing a non-empty response, then we pass its parameter to a generative model. The generative model synthesises a calorimeter response from the input particle parameters and a random noise vector. The schema of our method is visible on Fig. \ref{fig:method}.

\paragraph{Zero vs non-zero response classifier}
We use a binary neural network classifier to filter particles that do not produce any response of the ZDC. The model is a densely connected 3-layer network trained to distinguish between input parameters that produce results belonging to one of 2 classes - zero vs non-zero response. The first two layers of the network consist of 124 and 64 neurons respectively with ReLU activation functions. The last layer consists of a single neuron with a sigmoid activation function. The model was trained using binary cross-entropy loss.

\begin{figure}[h!]
\centering
\includegraphics[scale = 0.6]{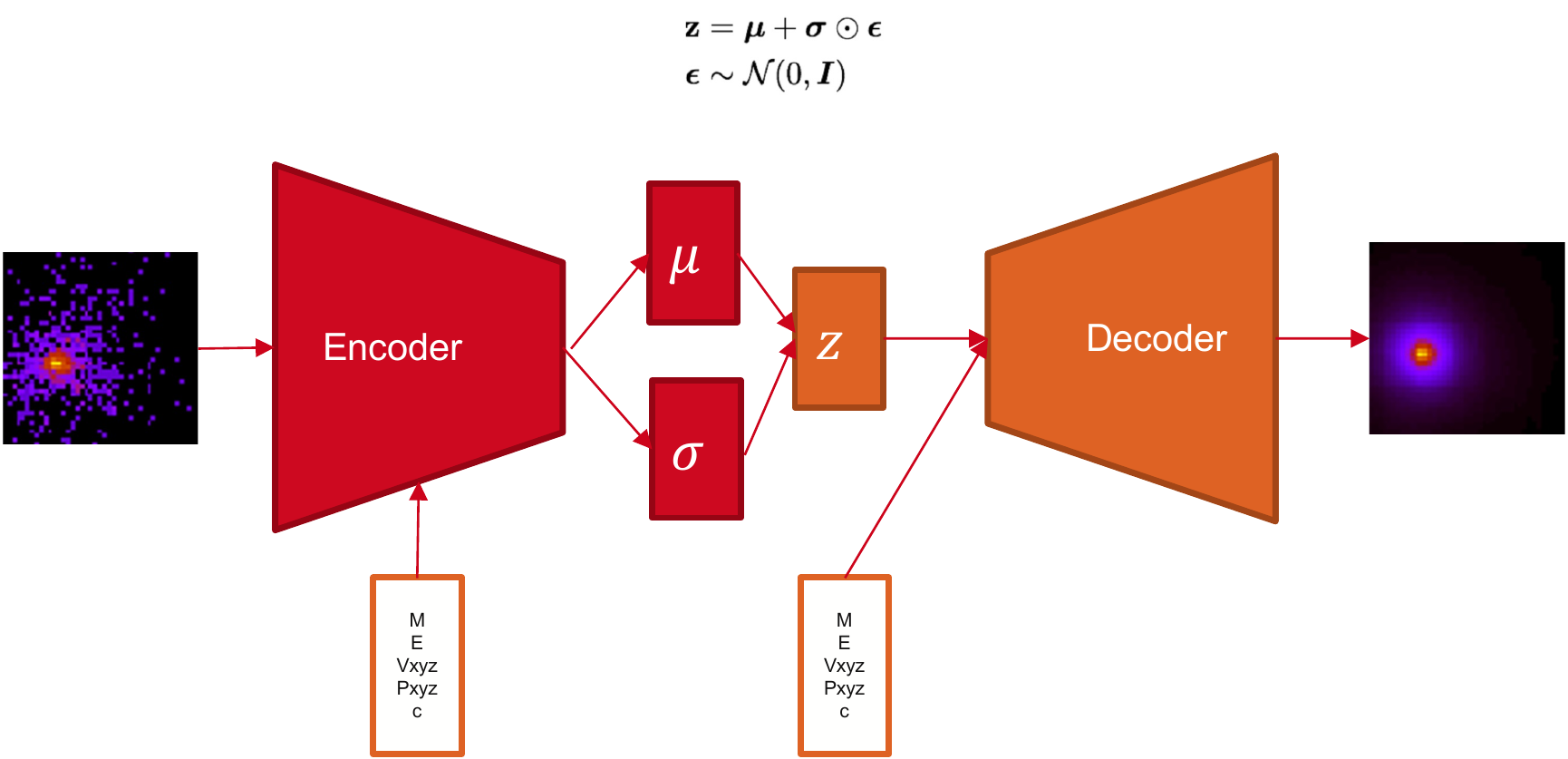}
\caption{ Variational autoencoder. }
\label{fig:vae}
\end{figure}

\paragraph{Variatonal Autoencoder}
As a baseline generative model, we apply a Variatonal Auteoncder \cite{kingma2013auto} to the problem of simulating data from the ZDC. The network consists of 2 parts: the encoder and the decoder. During training, the encoder compresses the data into a multivariate normal distribution of the latent variables. The decoder attempts to decompress the data and reconstruct the input. We use a conditional variant of the model, providing a conditional particle data vector for both the encoder and the decor, as presented in Fig.\ref{fig:vae}. For inference, only the encoder model is used, generating data from random normal noise and a conditional vector representing particle properties.

We use the following architecture for the encoder network: 
\begin{itemize}
    \setlength\itemsep{-0.2em}
    \item 3 convolutional layers with 32, 64 and 128 4x4 filters respectively, with stride = 2 and a LeakyReLU activation function
    \item a flattening layer to the output of which we concatenate corresponding conditional particle data vector
    \item a fully connected layer with 32 neurons and a LeakyReLU activation function
    \item two layers with 10 neurons for encoding mean and deviation of the latent variables
\end{itemize}

The decoder model has the following architecture:
\begin{itemize}
    \setlength\itemsep{-0.2em}
    \item an input layer of size 10 for latent variables to which we concatenate corresponding conditional particle data vector
    \item a fully connected layer with 4608 neurons, reshaped to a 6$\times$6$\times$128 shape
    \item  3 blocks each consisting of an upsampling layer, a convolutional layer (128/64/32 4$\times$4 filters), a batch normalization layer and a LeakyReLU activation function 
    \item an output convolutional layer with one 5$\times$5 filter and a ReLU activation function
\end{itemize}

\begin{figure}[h!]
\centering
\includegraphics[scale = 0.6]{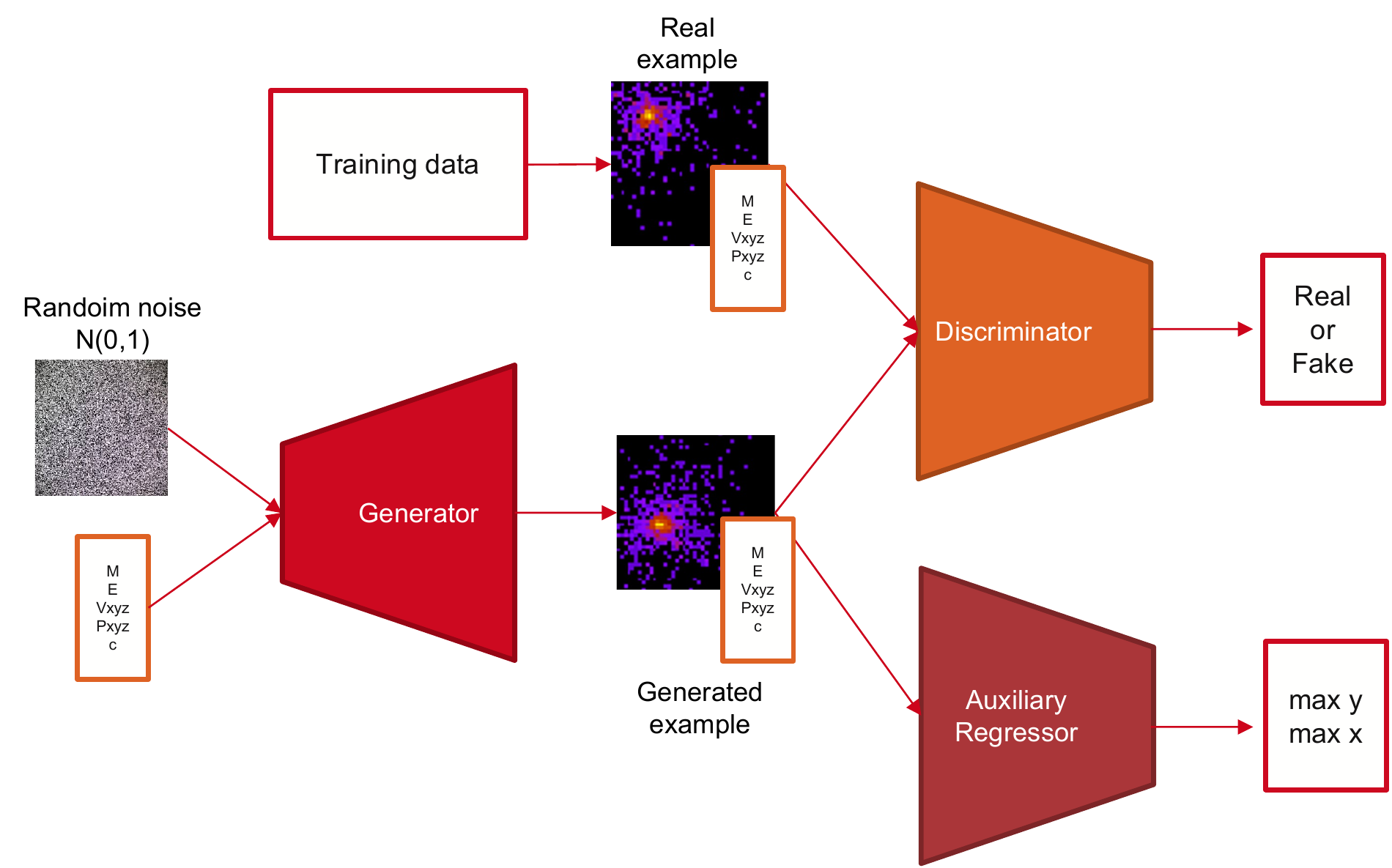}
\caption{ DC-GAN with auxiliary regressor. }
\label{fig:aux_reg}
\end{figure}

\paragraph{Deep Convolutional Generative Adversarial Network}
We  adopt the Deep Convolutional Generative Adversarial Network architecture introduced in \cite{radford2015unsupervised} as another baseline approach for the generative model. 
The GAN architecture consists of 2 networks: the generator and the discriminator. The generator learns to transform random noise into realistic data samples, while the discriminator learns to distinguish real and generated data. The two networks compete with each other during training. This process leads to a generator that can be used to generate new realistic data samples. We employ a conditional variant of the model, providing a conditional particle data vector for both the generator and the discriminator, as presented in Fig.\ref{fig:aux_reg}.

The generator model has the following architecture:
\begin{itemize}
    \setlength\itemsep{-0.2em}
    \item an input layer of size 10 for random normal noise to which we concatenate corresponding conditional particle data vector
    \item 2 fully connected layers with 256 and 21632 neurons, dropout=0.2 and LeakyReLU activation, reshaped to a 13$\times$13$\times$128 shape
    \item  2 blocks each consisting of an upsampling layer, a convolutional layer (128/64 3$\times$3 filters), a batch normalization layer and a LeakyReLU activation function 
    \item an output convolutional layer with one 3$\times$3 filter and a ReLU activation function
\end{itemize}

For the discriminator we use the following network:
\begin{itemize}
    \setlength\itemsep{-0.2em}
    \item 2 convolutional layers with 32, 16 4x4 filters respectively, with stride = 2 and LeakyReLU activation function
    \item a flattening layer to the output of which we concatenate corresponding conditional particle data vector
    \item 2 fully connected layer with 128 and 64 neurons with LeakyReLU activation function and dropout = 0.2
    \item an output layer with a single neuron with a sigmoid activation
\end{itemize}

\paragraph{Auxiliary regressor}
To improve how well the GAN model reflects the geometric properties of the real data we expand the standard GAN architecture by an auxiliary regressor. This additional network was pretrained to output the position coordinates of the maximum number of photons in the input image. The  regressor provides an additional source of loss to the generator by comparing the coordinates of the maximum of the generated examples with the maximum coordinates of the corresponding sample in the training set. During the training of the GAN model the loss coming from the auxiliary regressor is added to the loss of the generator.
The auxiliary regressor has the same architecture as the GAN discriminator, apart from the last layer with has 2 neurons with a ReLU activation function. The network was pre-trained on the same dataset used for training the generative model. The target variables corresponding to the coordinates of the maximum number of photons were calculated before the training as a data preprocessing step. The model was pretrained with mean squared error loss.

\paragraph{Postprocessing}
Contrary to many generative machine learning applications simulating High Energy Physics experiments provides an objective way to directly measure the quality of the generated samples. Motivated by this observation we introduce a simple, yet effective postprocessing step.

To improve the alignment of the distribution of generated samples and the original simulations we multiply the output of the generative model by a constant \textit{c}. Next, we calculate the Wasserstein distance between channels of the original and generative simulation on the training subset as shown in Sec. Results. We find the optimal value of \textit{c} by searching the parameter space between 0.9 to 1.1. We obtained the lowest Wasserstein distance for \textit{c} == 0.96 for GAN with the auxiliary regressor and \textit{c} == 1.03 for the standard GAN.

Additionally, we find a value for the standard deviation of the input noise vector that minimizes the Wasserstein distance between the original and the generative simulation. In our setup increasing the randomness of the input noise vectors helps to smooth the distribution of the generated results.  We achieve the lowest Wasserstein distance for \textit{N($\mu = 0, \sigma = 3$)}.

\section{Results}
\label{sec:results}
\label{sec:results:ev}

To evaluate the performance of the particle classifier we use well-established classification metrics - precision, recall, accuracy and F1- score. For generative models. the most common method for evaluation utilizes Frechet Inception Distance (FID) \cite{heusel2017gans}. However, to measure the quality of the simulation we propose a domain-specific evaluation scheme, as introduced in Sec. Zero Degree Calorimeter simulation. Following the calorimeter’s specification \cite{dellacasa1999alice} we base our evaluation procedure on 5 channels calculated from the pixels of generated images that correspond to the 5 optic fibre towers described in Sec. Zero Degree Calorimeter simulation. To measure the quality of the simulation we compare the distribution of channels for the original and generated data using Wasserstein distance \cite{tolstikhin2017wasserstein}. Those channels represent the physical properties of the simulated  particle showers and are used in the analysis of the calorimeter's output.

\begin{table}[h!]
\caption{Results for the binary particle classifier. We achieve high values for all evaluation metrics despite using a relatively simple neural network architecture.}
          \centering
          \begin{tabular}{l c c c c}
            \toprule
            &  precision $\uparrow$  &  recall $\uparrow$  &  F1 score $\uparrow$  &  accuracy $\uparrow$\\
            \midrule
            zero & 0.96 & 0.95 & 0.96 &0.95 \\
            non-zero & 0.93 & 0.95 & 0.94 & 0.95 \\
            \bottomrule
          \end{tabular}
          \label{tab:clf_results}
\end{table}

\begin{table}[h!]
\caption{: Results comparison for the HEP datasets. The VAE model performs better than GAN networks, even after introducing an auxiliary regressor to the GAN network. Applying the postprocessing step has a major impact on lowering the Wasserstein distance and allows GAN to outperform VAE. The GAN model with the auxiliary regressor and added postprocessing step achieves the lowest Wasserstein distance between channels calculated from the original and generated data.}
          \centering
          \begin{tabular}{l c }
            \toprule
               &  Wasserstein $\downarrow$  \\
            \midrule
            Real & - \\
            VAE & 6.45 \\
            DC-GAN & 8.25 \\
            DC-GAN + auxreg & 7.20 \\
            DC-GAN + postproc & 5.71 \\
            DC-GAN + auxreg+ postproc & 5.16 \\
            
            \bottomrule
          \end{tabular}
          \label{tab:all_results}
\end{table}

\begin{figure}[h!]
\centering
\begin{tabular}{c c }
     real data & \makecell{\includegraphics[scale = 0.4]{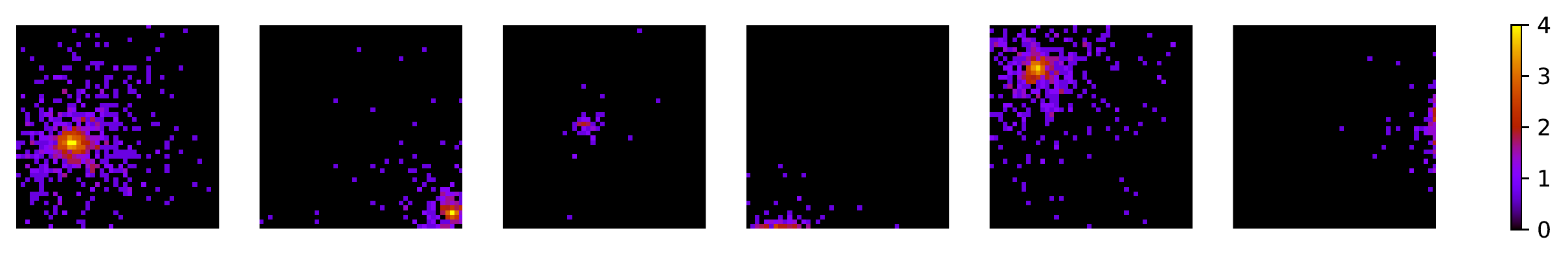}} \\
     VAE & \makecell{\includegraphics[scale = 0.4]{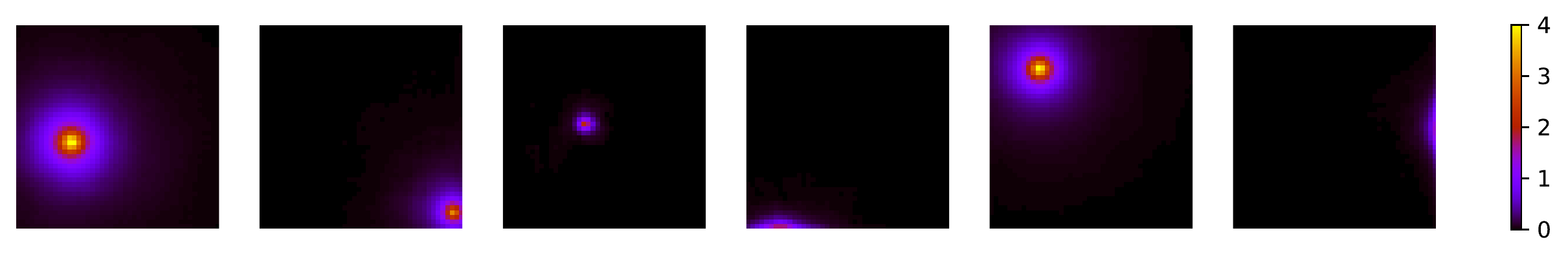}} \\
     DC-GAN & \makecell{\includegraphics[scale = 0.4]{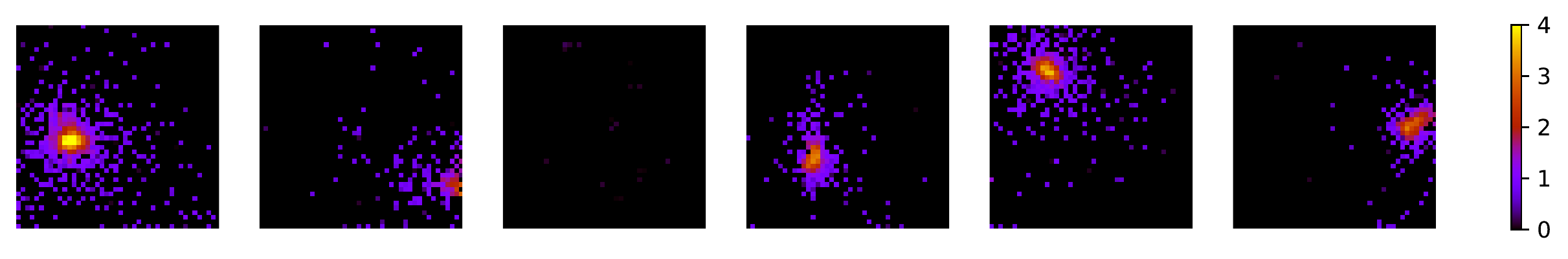}}\\
     \makecell{DC-GAN \\ + auxreg}  & \makecell{\includegraphics[scale = 0.4]{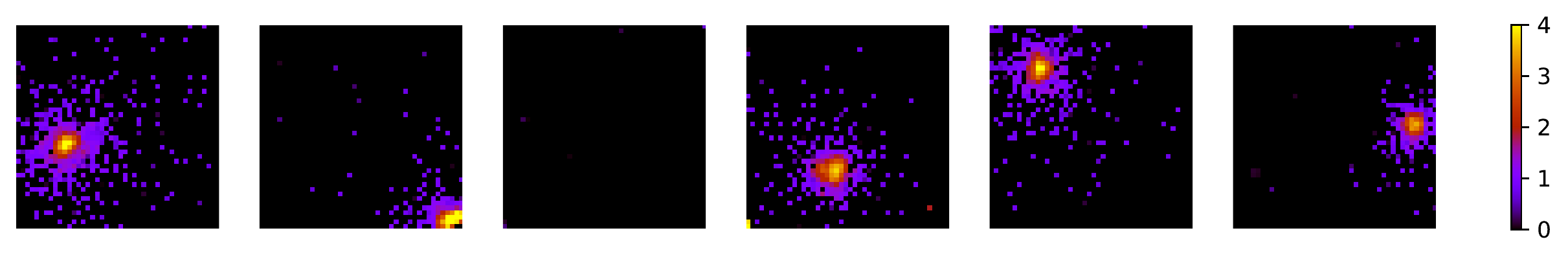}}\\
      \makecell{\textbf{DC-GAN} \\ \textbf{+ auxreg} \\ \textbf{+postproc}} & \makecell{\includegraphics[scale = 0.4]{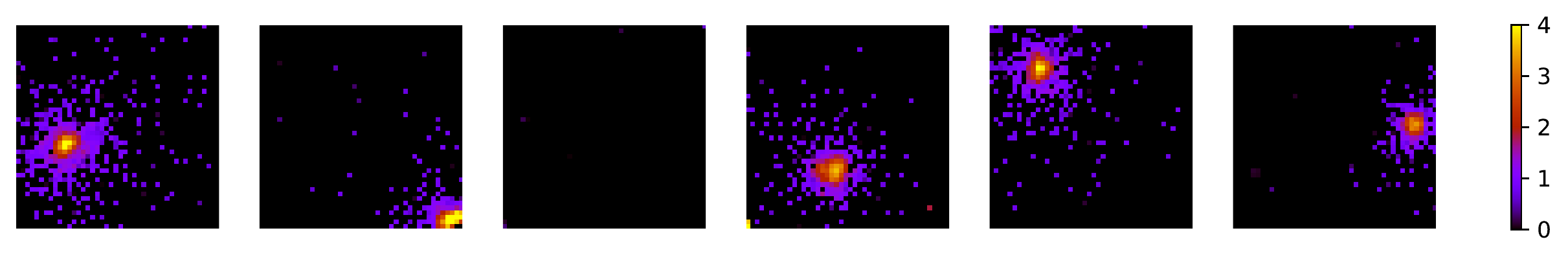}} \\
          
    \end{tabular}
\caption{Examples of calorimeter response simulations with different methods. Samples generated from VAE reproduce the exact locations of the particle showers, but they appear to be blurred. Although the results from DC-GAN are visually sound with the original data, the model was not able to properly capture relations from conditional values. Adding an auxiliary regressor to the GAN architecture improves the methods in terms of reconstructing the position of the shower's centre.
} 
\label{fig:HEP_examples}
\end{figure}

\begin{figure}[h!]
\centering
\begin{tabular}{c c c }
     &\textbf{Channel 1} & \textbf{Channel 2} \\
     VAE & \makecell{\includegraphics[scale = 0.25]{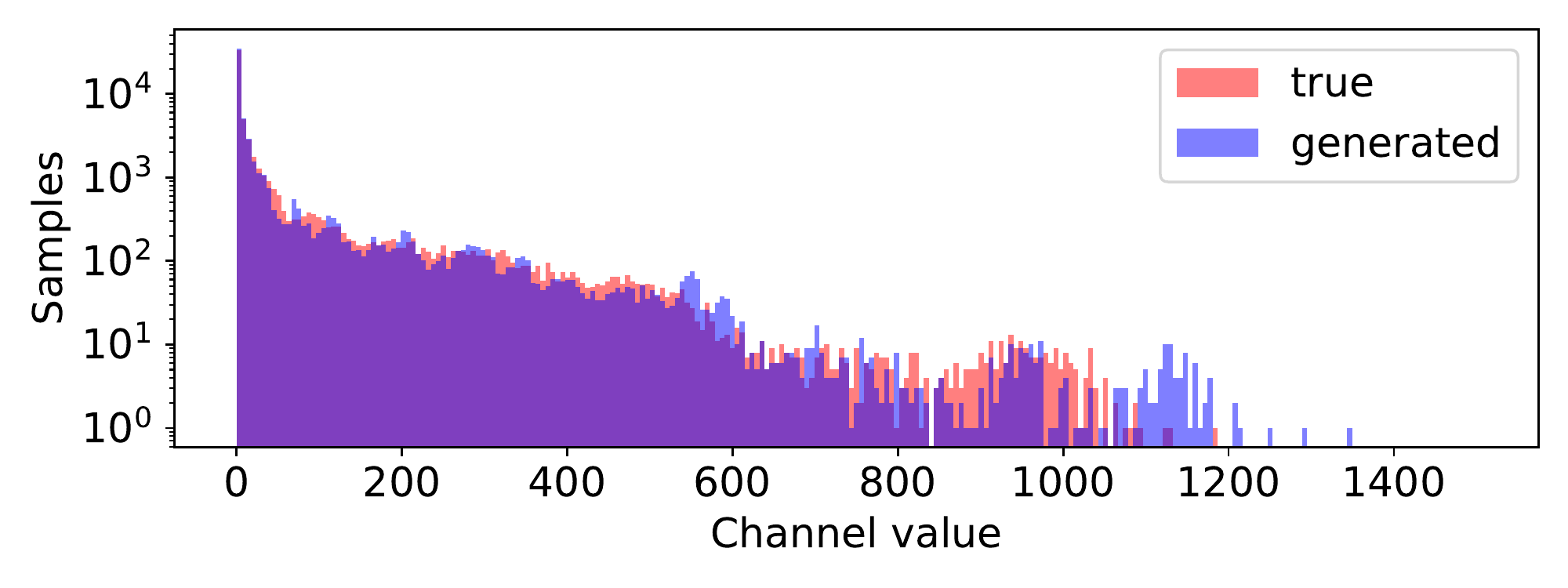}} & \makecell{\includegraphics[scale = 0.25]{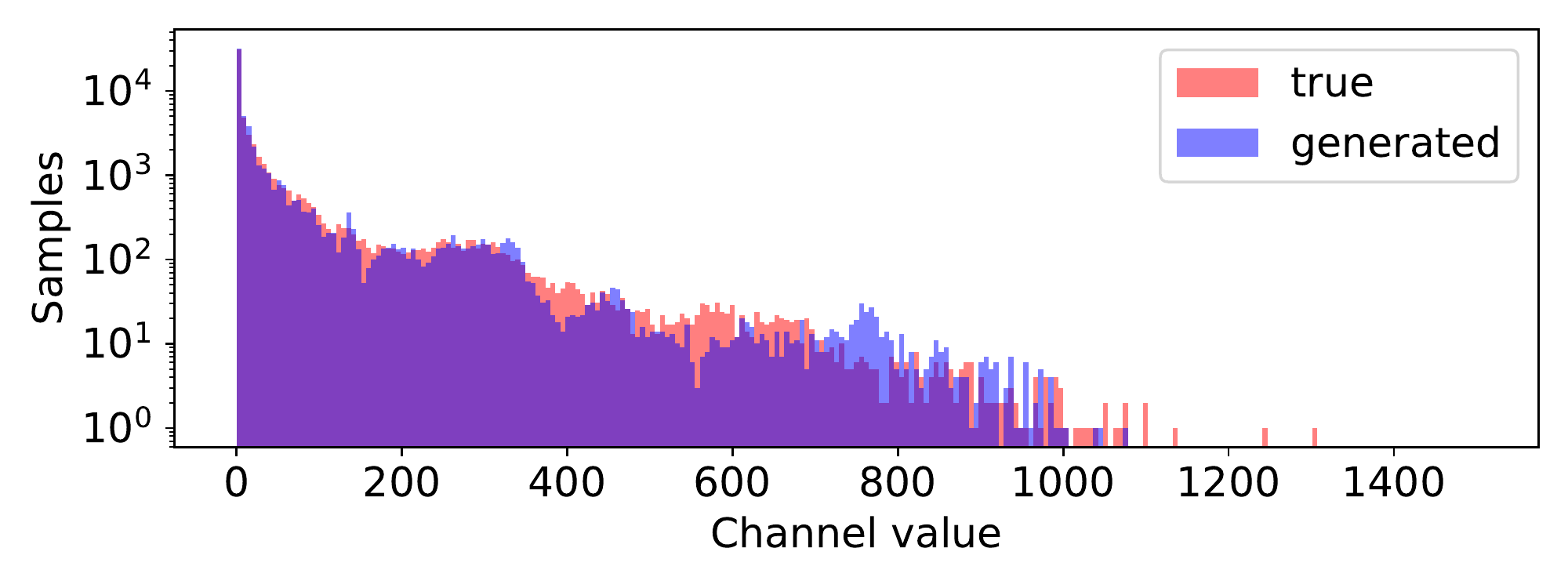}}\\
          DC-GAN & \makecell{\includegraphics[scale = 0.25]{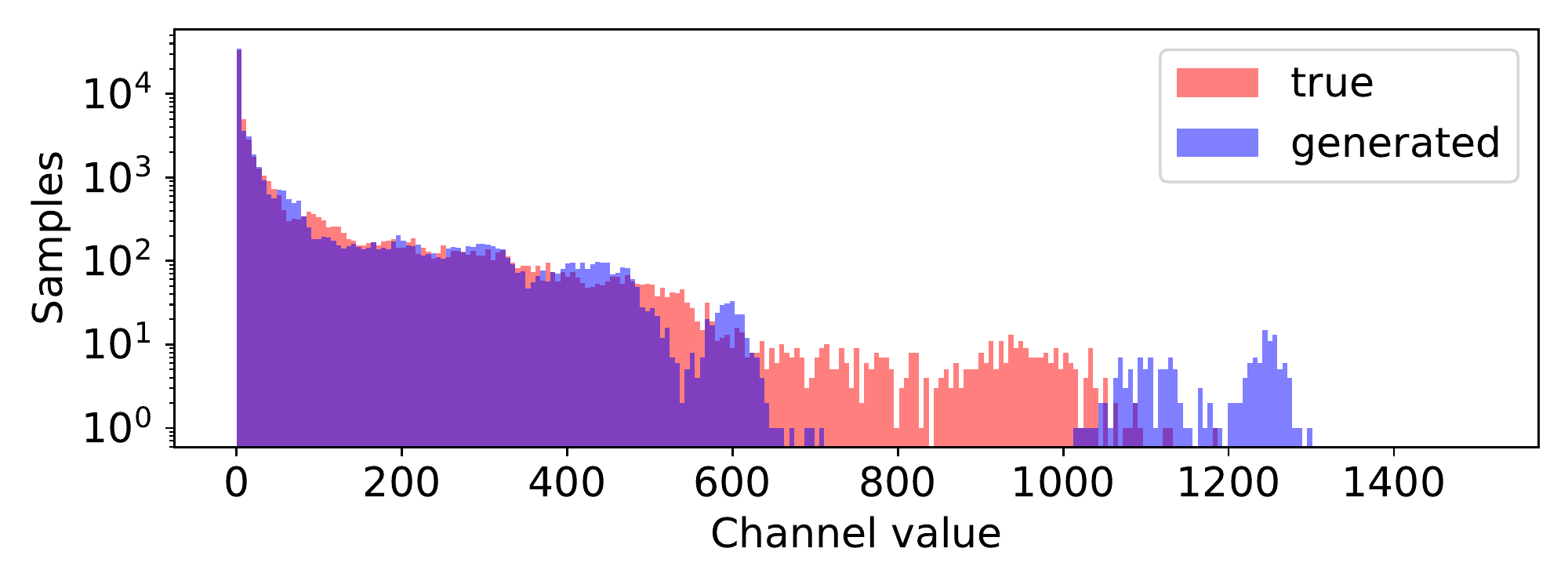}} & \makecell{\includegraphics[scale = 0.25]{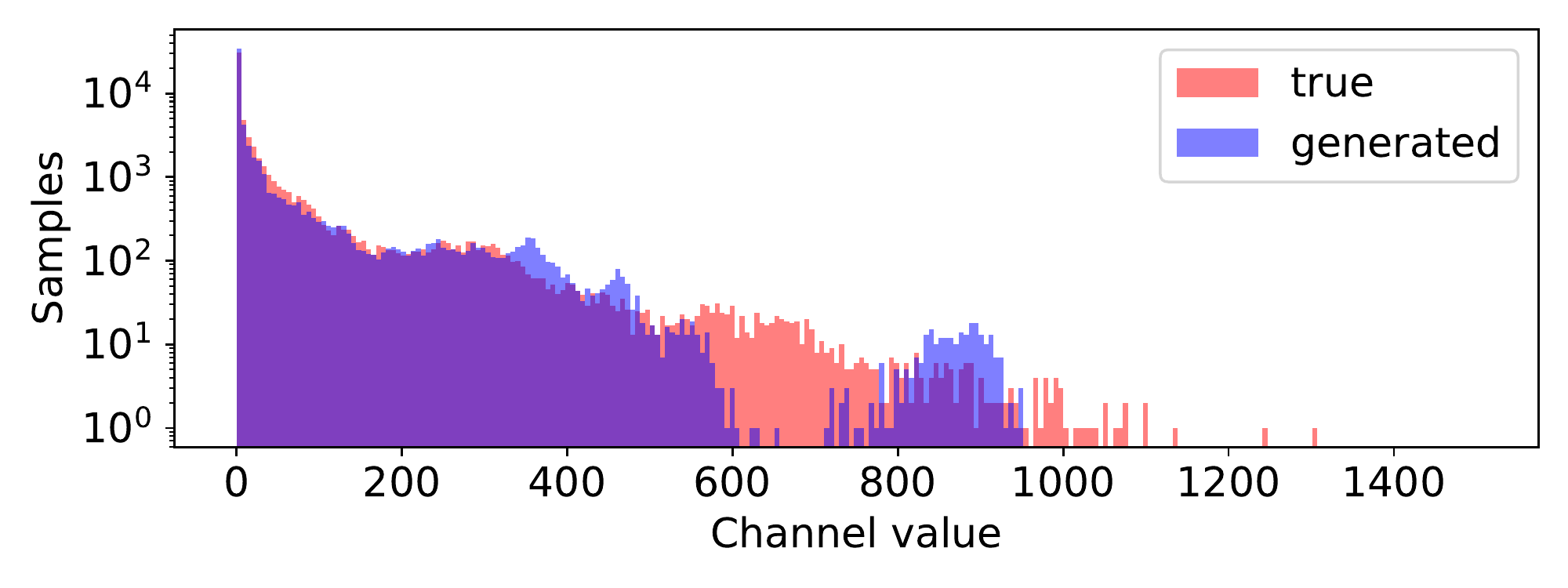}}\\
                \makecell{DC-GAN \\ + auxreg} & \makecell{\includegraphics[scale = 0.25]{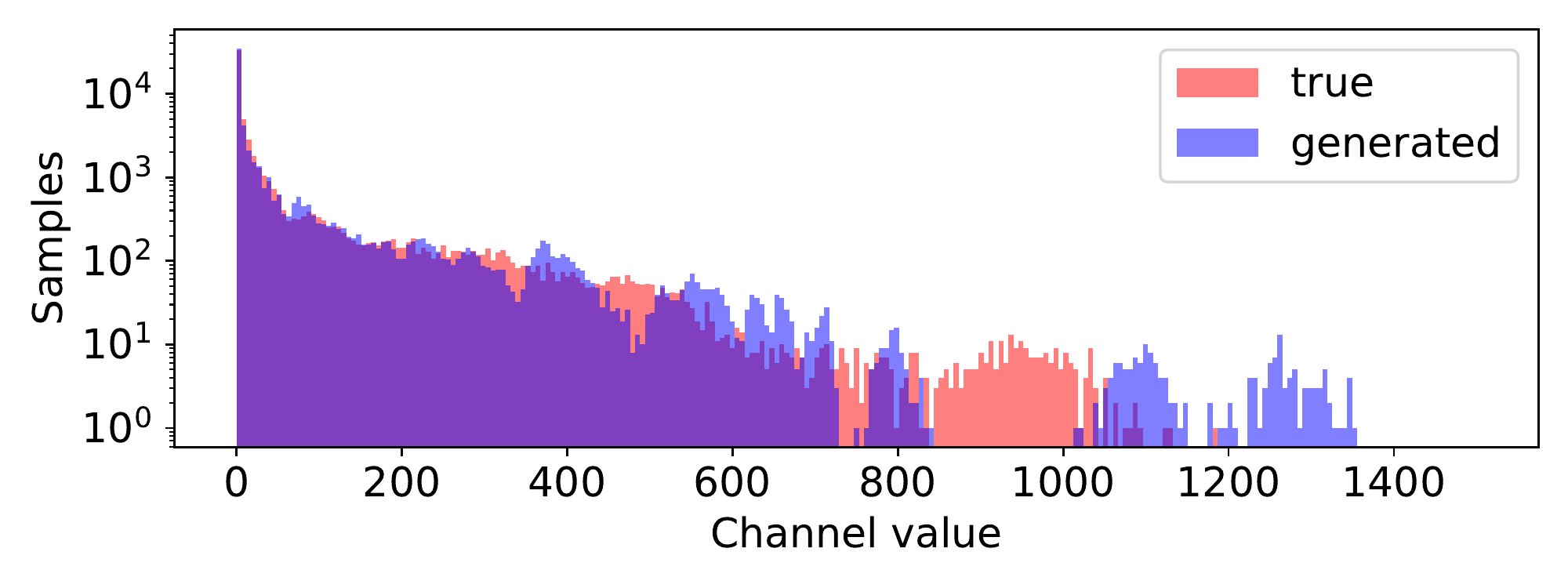}} & \makecell{\includegraphics[scale = 0.25]{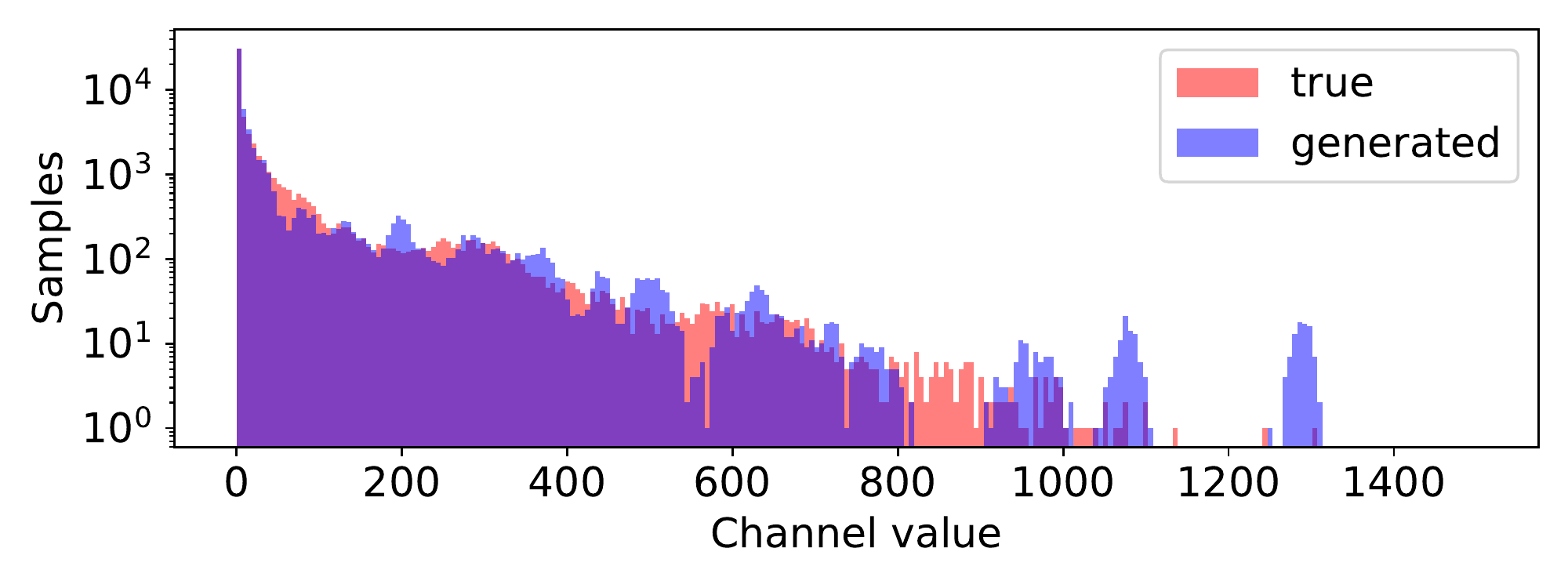}}\\
                     \makecell{DC-GAN \\ + postproc} & \makecell{\includegraphics[scale = 0.25]{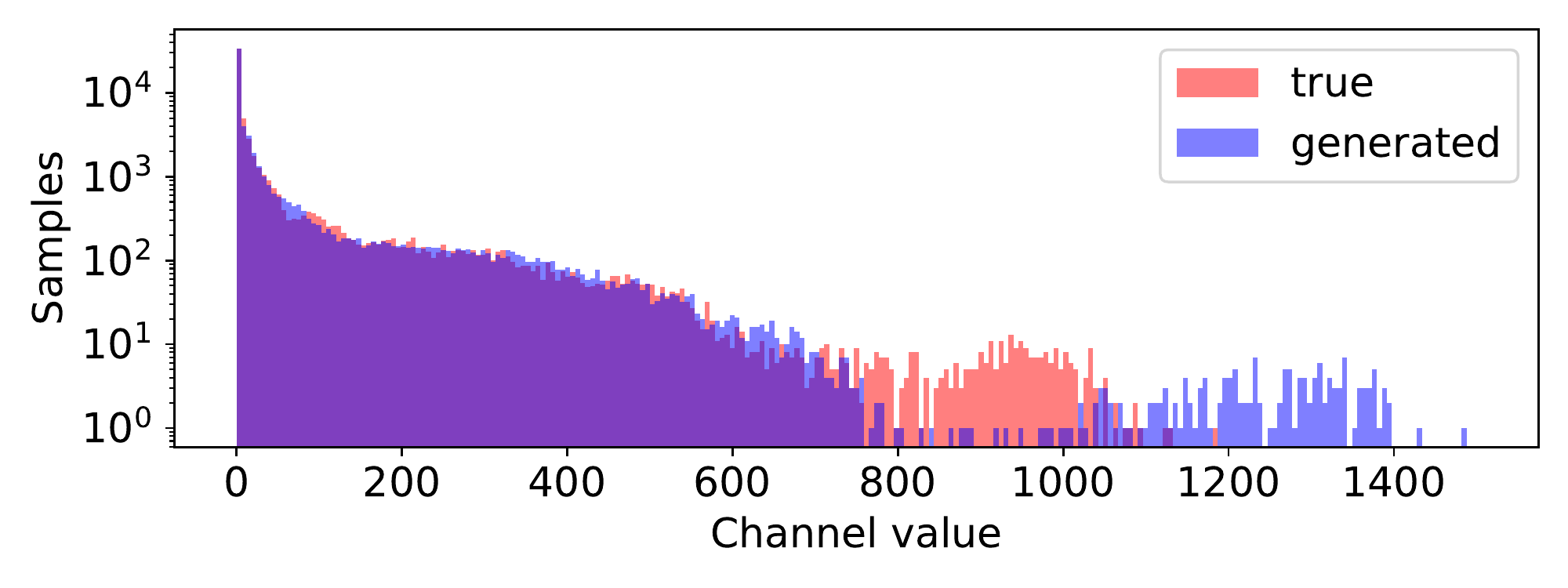}} & \makecell{\includegraphics[scale = 0.25]{figures/GAN+auxREG_ch1_12x4.pdf}}\\
      \makecell{\textbf{DC-GAN} \\ \textbf{+ auxreg} \\ \textbf{+ postproc}} & \makecell{\includegraphics[scale = 0.25]{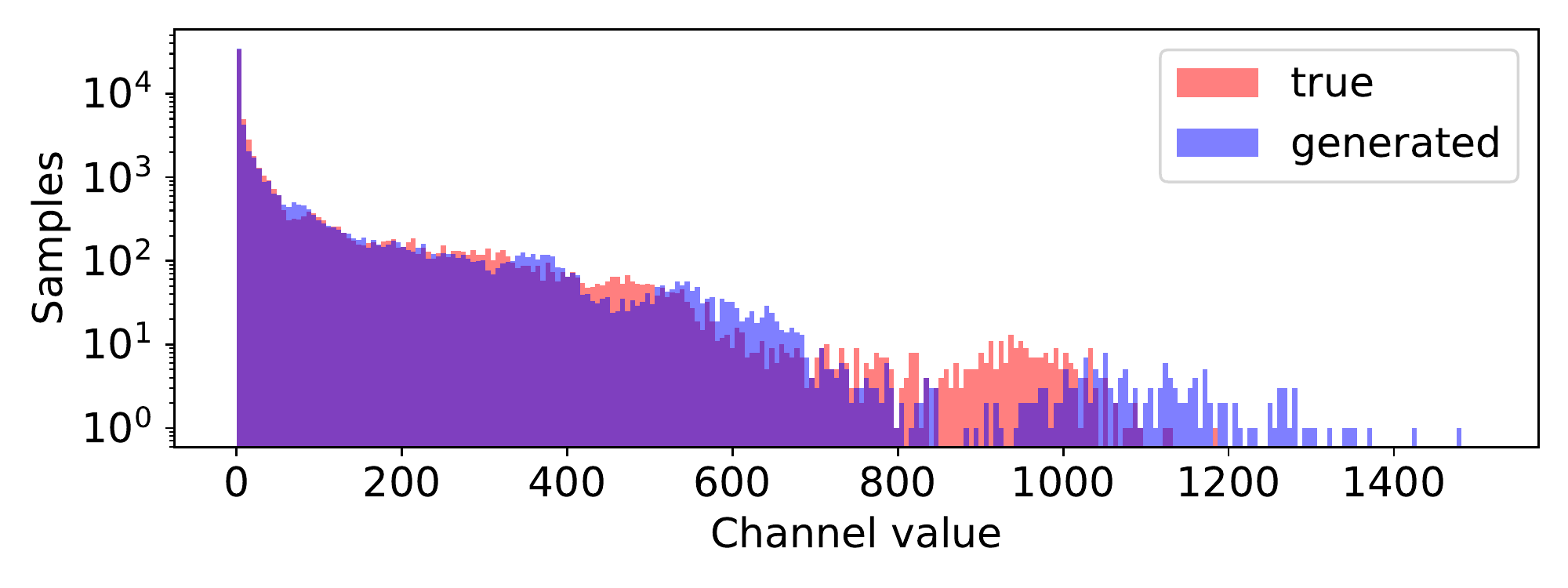}} & \makecell{\includegraphics[scale = 0.25]{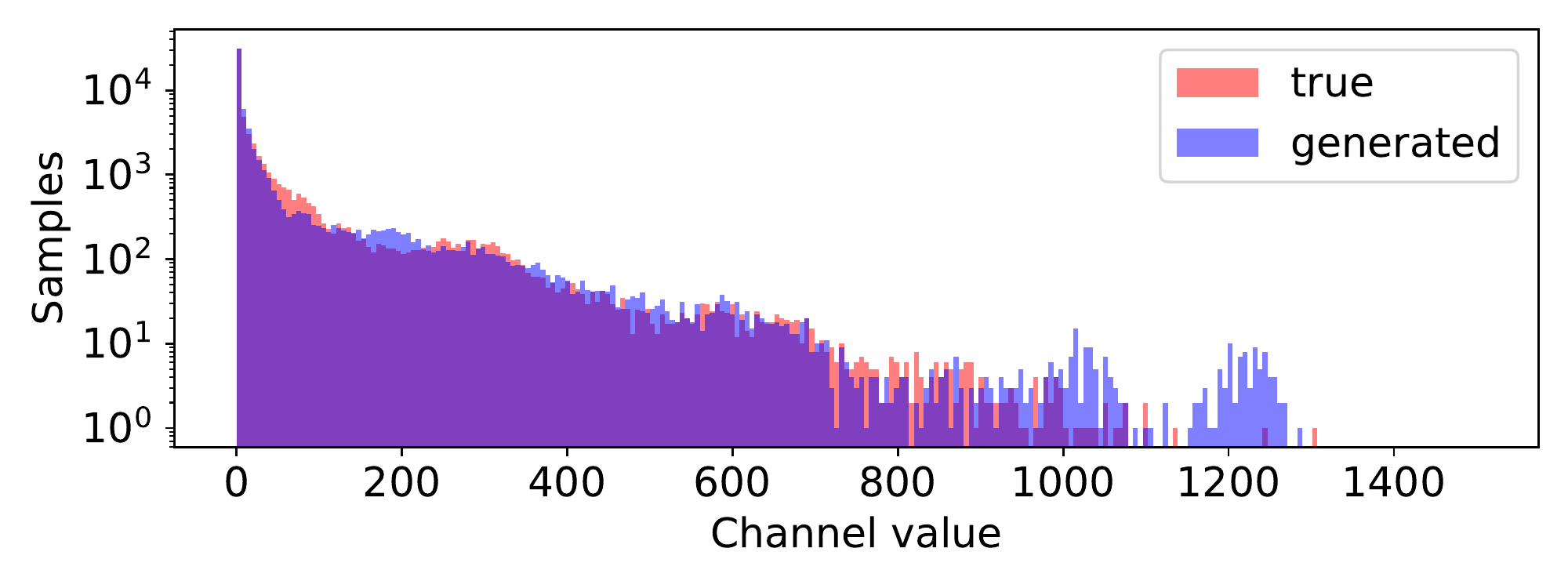}}\\
          
    \end{tabular}
\caption{ Comparison of channel values distribution for selected channels. VAE achieves good agreement between the two distributions. The GAN model is not able to cover the whole distribution of possible channel values and has trouble producing responses with high energy. Introducing the auxiliary does not fully mitigate this issue. However, adding the postprocessing step decreases the differences between the distribution of original and generated data and smooths the distribution of the synthesised results. }

\label{fig:ws_distributions}
\end{figure}

We present the results of our experiments using both qualitative and quantitative comparisons. In Tab.~\ref{tab:clf_results} we demonstrate that even using a relatively simple neural network we can successfully filter particles causing empty ZDC responses. In Fig.~\ref{fig:HEP_examples} we show examples of simulated calorimeter responses for different generative models. As demonstrated in Tab.~\ref{tab:all_results}, our approach outperforms other solutions on the ZDC dataset. Expanding the GAN architecture by an auxiliary regressor and introducing the postprocessing steps allows the GAN model to outperform VAE while producing visually sound calorimeter responses. The positive impact of this approach on the distribution of the generated samples is further confirmed by Fig.~\ref{fig:ws_distributions} where we compare channel distribution for all  competing approaches for 2 selected channels. Applying the postprocessing  increases the fidelity of the simulation by smoothing the distribution of generated responses and covering the whole range of possible outputs.

\section{Conclusions}
\label{sec:conlusions}

In this work, we apply generative machine learning models to the task of simulating the response of the Zero Degree Calorimeter from the ALICE experiment, CERN. We evaluate basic VAE and DC-GAN as baselines and apply two improvements to the GAN architecture.
Simulating HEP experiments requires the generative models to follow the strict physical properties of the modelled process. However, this domain also offers new possibilities to improve the performance of the generative models and provides objective evaluation metrics.
The developed generative machine learning models offer a cost-efficient alternative to Monte-Carlo-based simulations achieving a speed-up of 2 orders of magnitude.

\section*{Acknowledgments}
This research was funded by National Science Centre, Poland grant no 2020/39/O/ST6/01478, grant no 2018/31/N /ST6/02374 and grant no  2020/39/B/ST6/01511.

\clearpage

%%%%%%%%% REFERENCES
{\small
\bibliographystyle{splncs04.bst}
\bibliography{egbib.bib}
}
\end{document}